\documentclass[conference]{IEEEtran}
\IEEEoverridecommandlockouts
\usepackage{cite}
\usepackage{amsmath,amssymb,amsfonts}
\usepackage{algorithmic}
\usepackage{graphicx}
\usepackage{textcomp}
\usepackage{xcolor}
\def\BibTeX{{\rm B\kern-.05em{\sc i\kern-.025em b}\kern-.08em
    T\kern-.1667em\lower.7ex\hbox{E}\kern-.125emX}}
    
\usepackage{graphicx}
\usepackage{amsmath}
\usepackage{amsfonts}
\usepackage{booktabs}
\usepackage{ulem}
\usepackage{multirow}
\usepackage{bigstrut}
\usepackage[switch]{lineno}
\makeatletter
\newcommand\footnoteref[1]{\protected@xdef\@thefnmark{\ref{#1}}\@footnotemark}
\makeatother

\begin{document}

\title{Modeling Coverage for Non-Autoregressive \\Neural Machine Translation}

\author{
\IEEEauthorblockN{Yong Shan, Yang Feng$^\star$\thanks{$^\star$\ Yang Feng is the corresponding author.}, Chenze Shao}
\IEEEauthorblockA{\textit{Key Laboratory of Intelligent Information Processing}, \\ \textit{Institute of Computing Technology, Chinese Academy of Sciences (ICT/CAS)}, Beijing, China}
\IEEEauthorblockA{\textit{University of Chinese Academy of Sciences}, Beijing, China}
shanyong18s@ict.ac.cn,fengyang@ict.ac.cn,shaochenze18z@ict.ac.cn
}

\maketitle

\begin{abstract}
Non-Autoregressive Neural Machine Translation (NAT) has achieved significant inference speedup by generating all tokens simultaneously.
Despite its high efficiency, NAT usually suffers from two kinds of translation errors: \textit{over-translation} (e.g. repeated tokens) and \textit{under-translation} (e.g. missing translations), which eventually limits the translation quality.
In this paper, we argue that these issues of NAT can be addressed through coverage modeling, which has been proved to be useful in autoregressive decoding.
We propose a novel Coverage-NAT to model the coverage information directly by a token-level coverage iterative refinement mechanism and a sentence-level coverage agreement, which can remind the model if a source token has been translated or not and improve the semantics consistency between the translation and the source, respectively.
Experimental results on WMT14 En$\leftrightarrow$De and WMT16 En$\leftrightarrow$Ro translation tasks show that our method can alleviate those errors and achieve strong improvements over the baseline system. 
\end{abstract}


\section{Introduction}
Neural Machine Translation (NMT) has achieved impressive performance over the recent years \cite{cho2014learning, sutskever2014sequence, bahdanau2014neural, wu2016google, vaswani2017attention}. 
NMT models are typically built on the encoder-decoder framework where the encoder encodes the source sentence into distributed representations, and the decoder generates the target sentence from the representations in an autoregressive manner: the decoding at step $t$ is conditioned on the predictions of past steps. The autoregressive translation manner (AT) leads to high inference latency and restricts NMT applications.

Recent studies try to address this issue by replacing AT with non-autoregressive translation (NAT)  which generates all tokens simultaneously \cite{gu2017non}. 
However, the lack of sequential dependency between target tokens in NAT makes it difficult for the model to capture the highly multimodal distribution of the target translation. 
This phenomenon is termed as the \textit{multimodality} problem \cite{gu2017non}, which usually causes two kinds of translation errors during inference: \textit{over-translation} where the same token is generated repeatedly at consecutive time steps, and \textit{\textit{under-translation}} where the semantics of several source tokens are not fully translated \cite{wang2019non}.
Table \ref{tab:intro} shows a German-English example, where ``technical" and ``has" are translated repeatedly while ``also" is not translated at all.

\begin{table}[t!]
    \caption{A German-English example of \uline{\textit{over-translation}} and \uwave{\textit{under-translation}} errors in NAT. ``Src" and ``Ref" mean source and reference, respectively.}
    \centering
    \resizebox{\columnwidth}{!}{
        \begin{tabular}{ll}
            \toprule
            Src: & Das technische Personal hat mir ebenfalls viel gegeben. \\
            \midrule
            Ref: & The technical staff has also brought me a lot. \\
            NAT: & The \uline{technical technical} staff \uline{has has} \uwave{(also)} brought me a lot. \\
            \bottomrule
    \end{tabular}}
    \label{tab:intro}
\end{table}
Actually, there are \textit{over-translation} and \textit{under-translation} errors in AT models, too. They are usually addressed through explicitly modeling the coverage information step by step \cite{tu2016modeling, mi2016coverage}, where ``coverage" indicates whether a source token is translated or not and the decoding process is completed when all source tokens are ``covered".
Despite their successes in AT, existing NAT improvements to those two errors \cite{wang2019non,shao2019retrieving,shao2019minimizing} have no explorations in the coverage modeling.
It is difficult for NAT to model the coverage information due to the non-autoregressive nature, where step $t$ cannot know if a source token has been translated by other steps because all target tokens are generated simultaneously during decoding.

In this paper, we propose a novel Coverage-NAT to address this challenge by directly modeling coverage information in NAT. 
Our model works at both token level and sentence level. 
At token level, we propose a \textbf{t}oken-level \textbf{c}overage \textbf{i}terative \textbf{r}efinement mechanism (TCIR), which models the token-level coverage information with an iterative coverage layer and refines the source exploitation iteratively.
Specifically, we replace the top layer of the decoder with the iterative coverage layer and compute the coverage vector in parallel.
In each iteration, the coverage vector tracks the attention history of the previous iteration and then adjusts the attention of the current iteration, which encourages the model to concentrate more on the untranslated parts and hence reduces \textit{over-translation} and \textit{under-translation} errors.
In contrast to the step-wise manner \cite{tu2016modeling}, TCIR models the token-level coverage information in an iteration-wise manner, which does not hurt the non-autoregressive generation.
At sentence level, we propose a \textbf{s}entence-level \textbf{c}overage \textbf{a}greement (SCA) as a regularization term.
Inspired by the intuition that a good translation should ``cover" the source semantics at sentence level, SCA firstly computes the sentence representations of the translation and the source sentence, and then constrain the semantics agreement between them.
Thus, our model can improve the sentence-level coverage of translation, too.

We evaluate our model on four translation tasks (WMT14 En$\leftrightarrow$De, WMT16 En$\leftrightarrow$Ro). 
Experimental results show that our method outperforms the baseline system by up to +2.75 BLEU with competitive speedup (about 10$\times$), and further achieves near-Transformer performance on WMT16 dataset but 5.22$\times$ faster with the rescoring technique. 
The main contributions of our method are as follows:
\begin{itemize}
    \item To alleviate the \textit{over-translation} and \textit{under-translation} errors, we introduce the coverage modeling into NAT.
    \item We propose to model the coverage information at both token level and sentence level through iterative refinement and semantics agreement respectively, both of which are effective and achieve strong improvements.
\end{itemize}

\section{Background}
Recently, neural machine translation has achieved impressive improvements with the autoregressive decoding.
Despite its high performance, AT models suffer from long inference latency.
NAT \cite{gu2017non} is proposed to address this problem through parallelizing the decoding process. 

The NAT architecture can be formulated into an encoder-decoder framework \cite{sutskever2014sequence}. 
The same as AT models, the NAT encoder takes the embedding of source tokens as inputs and outputs the contextual source representations $\mathbf{E}_{enc}$ which are further used to compute the target length $T$ and inter-attention at decoder side.
\\ \noindent
\textbf{Multi-Head Attention:} The attention mechanism used in Transformer and NAT are multi-head attention ($\mathrm{Attn}$), which can be generally formulated as the weighted sum of value vectors $\mathbf{V}$ using query vectors $\mathbf{Q}$ and key vectors $\mathbf{K}$:
\begin{equation}
    \begin{split}
        \mathbf{A}=\mathrm{softmax}(\frac{\mathbf{Q} \mathbf{K}^T}{\sqrt{d_{model}}}) \\
        \mathrm{Attn}(\mathbf{Q}, \mathbf{K}, \mathbf{V}) = \mathbf{A} \mathbf{V}
    \end{split}
\end{equation}
where $d_{model}$ represents the dimension of hidden states. 
There are 3 kinds of multi-head attention in NAT: multi-head self attention, multi-head inter-attention and multi-head positional attention.
Refer to \cite{gu2017non} for details.
\\ \noindent
\textbf{Encoder:} A stack of $L$ identical layers. 
Each layer consists of a multi-head self attention and a position-wise feed-forward network ($\mathrm{FFN}$). 
$\mathrm{FFN}$ consists of two linear transformations with a ReLU activation \cite{vaswani2017attention}.
\begin{equation}
    \mathrm{FFN}(x)=\max(0,xW_1+b_1)W_2+b_2
\end{equation}
Note that we omit layer normalization, dropout and residual connections for simplicity.
Besides, a length predictor is used to predict the target length.
\\ \noindent
\textbf{Decoder:} A stack of $L$ identical layers. 
Each layer consists of a multi-head self attention, a multi-head inter-attention, a multi-head positional attention, and a position-wise feed-forward network. 

Given the input sentence in the source language $\mathbf{X}=\{x_1,x_2,\cdots,x_n\}$, NAT models generate a sentence in the target language $\mathbf{Y}=\{y_1,y_2,\cdots,y_T\}$ , where $n$ and $T$ are the lengths of the source and target sentence, respectively. The translation probability is defined as:
\begin{equation}
    P(\mathbf{Y}|\mathbf{X},\theta)=\prod_{t=1}^{T}p(y_t|\mathbf{X},\theta)
\end{equation}
The cross-entropy loss is applied to minimize the negative log-likelihood as:
\begin{equation}
    \mathcal{L}_{mle}(\theta)=-\sum_{t=1}^{T}log(p(y_t|\mathbf{X},\theta))
\end{equation}
During training, the target length $T$ is usually set as the reference length. During inference, $T$ is determined by the target length predictor, and then the translation is obtained by taking the token with the maximum likelihood at each step:
\begin{equation}
    \hat{y}_t=\arg \underset{y_t}{\max}\ p(y_t|\mathbf{X},\theta)
\end{equation}

\begin{figure*}[t!]
    \centering
    \includegraphics[width=0.85 \textwidth]{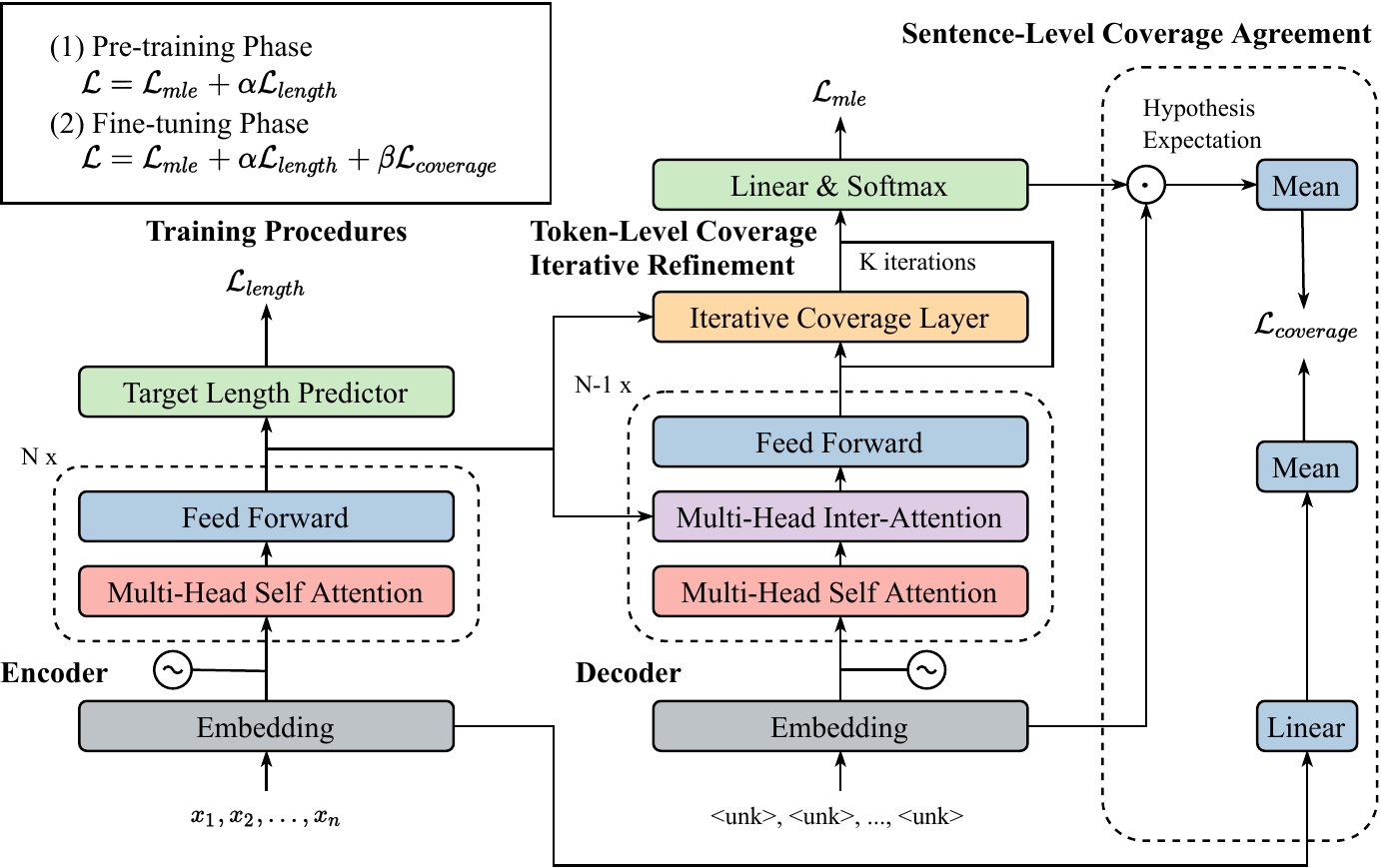}
    \label{fig:model}
    \caption{The architecture of Coverage-NAT. We propose the token-level coverage iterative refinement mechanism and the sentence-level coverage agreement to model the coverage information. At token level, we replace the top layer of the decoder with an iterative coverage layer which models the token-level coverage information and then refines the source exploitation in an iteration-wise manner. At sentence level, we maximize the source-semantics coverage of translation by constraining the sentence-level semantical agreement between the translation and the source sentence. We use a two-phase optimization to train our model. $\odot$ denotes the matrix multiplication.}
    \vspace{-10pt}
\end{figure*}

\section{Coverage-NAT}
In this section, we describe the details of Coverage-NAT. We first briefly introduce the NAT architecture in Section \ref{sec:arch}. Then, we illustrate the proposed token-level coverage iterative refinement (TCIR) and sentence-level coverage agreement (SCA) in Section \ref{sec:wordcover} and \ref{sec:sentcover}, respectively. Figure \ref{fig:model} shows the architecture of Coverage-NAT.
\subsection{NAT Architecture} \label{sec:arch}
The NAT architecture we used is similar to \cite{gu2017non} with the following differences in the decoder:
(1) Instead of copying source embedding according to fertility or uniform mapping, we feed a sequence of ``$\langle$unk$\rangle$" tokens to the NAT decoder directly. (2) We remove the positional attention in decoder layers. 
This design simplifies the decoder but works well in our experiments.

In Coverage-NAT, the bottom $L-1$ layers of the decoder are identical. Specifically, we design a novel iterative coverage layer as the top layer, which refines the source exploitations by iteratively modeling token-level coverage.
\subsection{Token-Level Coverage Iterative Refinement} \label{sec:wordcover}
The token-level coverage information has been applied to phrase-based SMT \cite{koehn2003statistical} and AT \cite{tu2016modeling}, where the decoder maintains a coverage vector to indicate whether a source token is translated or not and then pays more attention to the untranslated parts.
Benefit from the autoregressive nature, AT models can easily model the coverage information step by step.
Generally, it is a step-wise refinement of source exploitation with the assistance of coverage vector.
However, it encounters difficulties in non-autoregressive decoding. 
Since all target tokens are generated simultaneously, step $t$ cannot know if a source token has been translated by other steps during decoding.

The key to solving this problem of NAT lies in: (1) How to compute the coverage vector under the non-autoregressive setting? (2) How to use the coverage vector to refine the source exploitation under the non-autoregressive setting?
To address it, we propose a novel token-level coverage iterative refinement mechanism. Our basic idea is to replace the top layer of the decoder with an iterative coverage layer, where we model the coverage information to refine the source exploitation in an iteration-wise manner. The iterative coverage layer also contains a multi-head self attention, a multi-head inter-attention and an $\mathrm{FFN}$.

For question (1), we maintain an attention precedence which means the former steps have priorities to attend to source tokens and later steps shouldn't scramble for those already ``covered" parts. We manually calculate the coverage vector $\mathbf{C}^k$ on the basis of the inter-attention weights of the previous iteration. At step $t$, we compute the accumulation of past steps' attention on $x_i$ as $C_{t,i}^k$ to represent the coverage of $x_i$:
\begin{equation}
    C_{t,i}^{k}=\min(\sum_{t'=0}^{t-1}A_{t',i}^{k-1},1) \label{eqn:coverage}
\end{equation}
where $k$ is the iteration number, $A_{t',i}^{k-1}$ is the inter-attention weights from step $t'$ to $x_i$ in the previous iteration. We use $\min$ function to ensure all coverage do not to exceed 1.

For question (2), we use $\mathbf{C}^k$ as a bias of inter-attention so that step $t$ will pay more attention to source tokens untranslated by past steps in the previous iteration. 
A full iteration can be given by:
\begin{eqnarray}
    \mathbf{\hat{H}}^{k}&=&\mathrm{Attn}(\mathbf{H}^{k-1},\mathbf{H}^{k-1},\mathbf{H}^{k-1}) \label{eqn:selfattn} \\
    \mathbf{B}^{k}&=&(1-\mathbf{C}^k) \label{eqn:bias} \\
    \mathbf{A}^{k}&=&\mathrm{softmax}(\frac{\mathbf{\hat{H}}^{k}\mathbf{E}_{enc}^T}{\sqrt{d_{model}}}+\lambda \times \mathbf{B}^{k}) \label{eqn:addbias} \\
    \mathbf{H}^{k}&=&\mathrm{FFN}(\mathbf{A}^{k} \mathbf{E}_{enc}) \label{eqn:iterffn}
\end{eqnarray}
where $\mathbf{H}^{k}$ is the hidden states of iteration $k$, $\lambda \in \mathbb{R}^1$ is trainable and is initialized to 1, $\times$ means the scalar multiplication.
In Equation \ref{eqn:selfattn}, we compute the self attention and store the output as $\mathbf{\hat{H}}^{k}$.  
In Equation \ref{eqn:bias}-\ref{eqn:addbias}, we firstly calculate the subtraction between 1 and $\mathbf{C}^k$ to make the source tokens with a lower coverage obtain a higher bias, and then integrate the bias $\mathbf{B}^{k}$ into the inter-attention weights by a trainable scaling factor $\lambda$. 
Finally, we output the hidden states of iteration $k$ in Equation \ref{eqn:iterffn}. 
In this way, each step is discouraged to consider source tokens heavily attended by past steps in the previous iteration, and pushed to concentrate on source tokens less attended.
The source exploitation is refined iteratively with token-level coverage, too.

We use the hidden states and the inter-attention weights of $(L-1)$th layer as $\mathbf{H}^0$ and $\mathbf{A}^0$, respectively.
After $K$ iterations, we use $\mathbf{H}^K$ as the output of the decoder.
\subsection{Sentence-Level Coverage Agreement} \label{sec:sentcover}
In AT models, the sentence-level agreement between the source side and the target side has been exploited by Yang et al. \cite{yang2019sentence}, which constrains the agreement between source sentence and reference sentence.
However, it can only enhance the representations of the source side and target side instead of improving the semantics consistency between the source sentence and the translation.

Different from them, we consider the sentence-level agreement from a coverage perspective.
Intuitively, a translation ``covers" source semantics at sentence level if the representation of translation is as close as possible to the source sentence. 
Thus, we propose to constrain the sentence-level agreement between the source sentence and the translation.
For the source sentence, we compute the sentence representation by considering the whole sentence as a bag-of-words. 
Specifically, we firstly map the source embeddings to target latent space by a linear transformation, and then use a mean pooling to obtain the sentence representation $\bar{\mathbf{E}}_{src}$:
\begin{equation}
    \bar{\mathbf{E}}_{src}=\mathrm{Mean}(\mathrm{ReLU}(\mathbf{E}_{src}\mathbf{W}_{s})) \label{eqn:srcbag}
\end{equation}
where $\mathbf{E}_{src}=\{\mathbf{E}_{src,1},\cdots,\mathbf{E}_{src,n}\}$ is the word embeddings of source sentence, $\mathbf{W}_{s} \in 
\mathbb{R}^{d_{model} \times d_{model}}$ represents the parameters of linear transformation and $\mathrm{ReLU}$ is the activation function.
For the translation, we need to calculate the expectation of hypothesis at each position, which takes all possible translation results into consideration.
Although it is inconvenient for AT to calculate the expectation by multinomial sampling \cite{chatterjee2010minimum}, we can do it easily in NAT due to the token-independence assumption:
\begin{equation}
    \mathbb{E}_{hyp,t}=\mathbf{p}_{t} \mathbf{W}_{e}
\end{equation}
where $\mathbf{p}_{t} \in \mathbb{R}^{d_{vocab}}$ is the probability vector of hypothesis at step $t$, $\mathbf{W}_{e} \in \mathbb{R}^{d_{vocab} \times d_{model}}$ is the weights of target-side word embedding. Then, we calculate the sentence representation of translation by mean pooling as follows:
\begin{equation}
    \bar{\mathbf{E}}_{hyp}=\mathrm{Mean}(\{\mathbb{E}_{hyp,1},\cdots,\mathbb{E}_{hyp,T}\})
\end{equation}
Finally, we compute the $\mathrm{L2}$ distance between the source sentence and the translation as the sentence-level coverage agreement. Note that we use a scaling factor $\frac{1}{\sqrt{d_{model}}}$ to stablize the training:
\begin{equation}
    \mathcal{L}_{coverage}=\frac{\mathrm{L2}(\bar{\mathbf{E}}_{src}, \bar{\mathbf{E}}_{hyp})}{\sqrt{d_{model}}}
\end{equation}
\subsection{Two-Phase Optimization} \label{sec:optim}
In our approach, we firstly pre-train the proposed model to convergence with basic loss functions, and then fine-tune the proposed model with the SCA and basic loss functions.
\\ \noindent
\textbf{Pre-training} \, During the pre-training phase, we optimize the basic loss functions for NAT:
\begin{equation}
    \mathcal{L}=\mathcal{L}_{mle}+\alpha \mathcal{L}_{length}
\end{equation}
where $\mathcal{L}_{length}$ is the loss of target length predictor \cite{gu2017non}. $\alpha$ is a hyper-parameter to control the weight of $\mathcal{L}_{length}$.
\\ \noindent
\textbf{Fine-tuning} \, During the fine-tuning phase, we optimize the SCA and basic loss functions jointly:
\begin{equation}
    \mathcal{L}=\mathcal{L}_{mle}+\alpha \mathcal{L}_{length}+\beta \mathcal{L}_{coverage}
\end{equation}
where $\beta$ is a hyper-parameter to control the weight of $\mathcal{L}_{coverage}$. 
We have tried to train the model with all loss functions jointly from scratch. 
The result is worse than fine-tuning. 
We think that this is because the SCA will interfere the optimization of translation when the training begins.
\section{Experiments Setup}
\begin{table*}[t!]
    \caption{The BLEU (\%) and the speedup on test sets of WMT14 En$\leftrightarrow$De, WMT16 En$\leftrightarrow$Ro tasks.}
    \centering
    \resizebox{\textwidth}{!}{%
        \begin{tabular}{lcccccc}
            \hline
            \multicolumn{1}{l|}{\multirow{2}{*}{Models}}            & \multicolumn{1}{c|}{WMT14}             & \multicolumn{1}{c|}{WMT14}             & \multicolumn{1}{c|}{WMT16}             & \multicolumn{1}{c|}{WMT16}             & \multicolumn{1}{c|}{\multirow{2}{*}{Latency}} & \multirow{2}{*}{Speedup} \\
            \multicolumn{1}{l|}{}                                   & \multicolumn{1}{c|}{En$\rightarrow$De} & \multicolumn{1}{c|}{De$\rightarrow$En} & \multicolumn{1}{c|}{En$\rightarrow$Ro} & \multicolumn{1}{c|}{Ro$\rightarrow$En} & \multicolumn{1}{c|}{}                         &                          \\ \hline
            \multicolumn{1}{l|}{Transformer \cite{vaswani2017attention}}                        & \multicolumn{1}{c|}{27.19}             & \multicolumn{1}{c|}{31.25}             & \multicolumn{1}{c|}{32.80}              & \multicolumn{1}{c|}{32.61}             & \multicolumn{1}{c|}{276.88 ms}                & 1.00$\times$               \\ \hline
            \textit{Existing NAT Implementations}                                  &                                        &                                        &                                        &                                        &                                               &                          \\ \hline
            \multicolumn{1}{l|}{NAT-FT \cite{gu2017non}}                             & \multicolumn{1}{c|}{17.69}             & \multicolumn{1}{c|}{21.47}             & \multicolumn{1}{c|}{27.29}             & \multicolumn{1}{c|}{29.06}             & \multicolumn{1}{c|}{/}                        & 15.60$\times$            \\
            \multicolumn{1}{l|}{NAT-FT (rescoring 10) \cite{gu2017non}}              & \multicolumn{1}{c|}{18.66}             & \multicolumn{1}{c|}{22.41}             & \multicolumn{1}{c|}{29.02}             & \multicolumn{1}{c|}{30.76}             & \multicolumn{1}{c|}{/}                        & 7.68$\times$            \\
            \multicolumn{1}{l|}{NAT-Base \footnoteref{footnote:fairseq}}                           & \multicolumn{1}{c|}{18.60}              & \multicolumn{1}{c|}{22.70}              & \multicolumn{1}{c|}{28.78}             & \multicolumn{1}{c|}{29.00}                & \multicolumn{1}{c|}{19.36 ms}                 & 14.30$\times$            \\
            \multicolumn{1}{l|}{NAT-IR ($i_{dec}=5$) \cite{lee2018deterministic}}              & \multicolumn{1}{c|}{20.26}             & \multicolumn{1}{c|}{23.86}             & \multicolumn{1}{c|}{28.86}             & \multicolumn{1}{c|}{29.72}             & \multicolumn{1}{c|}{/}                        & 3.11$\times$            \\
            \multicolumn{1}{l|}{Reinforce-NAT \cite{shao2019retrieving}}                      & \multicolumn{1}{c|}{19.15}             & \multicolumn{1}{c|}{22.52}             & \multicolumn{1}{c|}{27.09}             & \multicolumn{1}{c|}{27.93}             & \multicolumn{1}{c|}{/}                        & 10.73$\times$           \\
            \multicolumn{1}{l|}{NAT-REG \cite{wang2019non}}                            & \multicolumn{1}{c|}{20.65}             & \multicolumn{1}{c|}{24.77}             & \multicolumn{1}{c|}{/}                 & \multicolumn{1}{c|}{/}                 & \multicolumn{1}{c|}{/}                        & 27.60$\times$            \\
            \multicolumn{1}{l|}{BoN-NAT \cite{shao2019minimizing}}                            & \multicolumn{1}{c|}{20.90}              & \multicolumn{1}{c|}{24.61}             & \multicolumn{1}{c|}{28.31}             & \multicolumn{1}{c|}{29.29}             & \multicolumn{1}{c|}{/}                        & 10.77$\times$           \\
            \multicolumn{1}{l|}{Hint-NAT \cite{li2019hint}}                           & \multicolumn{1}{c|}{21.11}             & \multicolumn{1}{c|}{25.24}             & \multicolumn{1}{c|}{/}                 & \multicolumn{1}{c|}{/}                 & \multicolumn{1}{c|}{/}                        & 30.20$\times$           \\
            \multicolumn{1}{l|}{FlowSeq-large \cite{ma2019flowseq}}                      & \multicolumn{1}{c|}{23.72}             & \multicolumn{1}{c|}{28.39}             & \multicolumn{1}{c|}{29.73}             & \multicolumn{1}{c|}{30.72}             & \multicolumn{1}{c|}{/}                        & /                        \\ \hline
            \textit{Our NAT Implementations}                                  &                                        &                                        &                                        &                                        &                                               &                          \\ \hline
            \multicolumn{1}{l|}{Coverage-NAT}              & \multicolumn{1}{c|}{21.35}             & \multicolumn{1}{c|}{25.04}             & \multicolumn{1}{c|}{30.05}             & \multicolumn{1}{c|}{30.33}              & \multicolumn{1}{c|}{27.58 ms}                 & 10.04$\times$           \\
            \multicolumn{1}{l|}{Coverage-NAT (rescoring 9)} & \multicolumn{1}{c|}{23.76}              & \multicolumn{1}{c|}{27.84}             & \multicolumn{1}{c|}{31.93}             & \multicolumn{1}{c|}{32.32}             & \multicolumn{1}{c|}{53.08 ms}                 & 5.22$\times$            \\ \hline
        \end{tabular}%
    }
    \vspace{-10pt}
    \label{tab:main-results}
\end{table*}
\noindent\textbf{Datasets} \, 
We evaluate the effectiveness of our proposed method on WMT14 En$\leftrightarrow$De (4.5M pairs) and WMT16 En$\leftrightarrow$Ro (610k pairs) datasets.
For WMT14 En$\leftrightarrow$De, we employ newstest2013 and newstest2014 as development and test sets.
For WMT16 En$\leftrightarrow$Ro, we employ newsdev2016 and newstest2016 as development and test sets. 
We tokenize all sentences into subword units \cite{sennrich2016neural} with 32k merge operations. 
We share the vocabulary between the source and target sides.
\\ \noindent
\textbf{Metrics} \, We evaluate the model performance with case-sensitive BLEU \cite{papineni2002bleu} for the En-De dataset and case-insensitive BLEU for the En-Ro dataset. 
Latency is computed as the average of per sentence decoding time (ms) on the full test set of WMT14 En$\rightarrow$De without mini-batching. 
We test latency on 1 Nvidia RTX 2080Ti GPU.
\\ \noindent
\textbf{Baselines} \, We take the Transformer model \cite{vaswani2017attention} as our AT baseline as well as the teacher model. We choose the following models as our NAT baselines:
(1) \textbf{NAT-FT} which equips vanilla NAT with fertility fine-tuning \cite{gu2017non}.
(2) \textbf{NAT-Base} which is the NAT implementation of Fairseq\footnote{\label{footnote:fairseq}https://github.com/pytorch/fairseq} which removes the multi-head positional attention and feeds the decoder with a sequence of ``$\langle$unk$\rangle$" instead of copying the source embedding.
(3) \textbf{NAT-IR} which iteratively refines the translation for multiple times \cite{lee2018deterministic}.
(4) \textbf{Reinforce-NAT} which introduces a sequence-level objective using reinforcement learning \cite{shao2019retrieving}.
(5) \textbf{NAT-REG} which proposes two regularization terms to alleviate translation errors \cite{wang2019non}.
(6) \textbf{BoN-NAT} which minimizes the bag-of-ngrams differences between the translation and the reference\cite{shao2019minimizing}.
(7) \textbf{Hint-NAT} which leverages the hints from AT models to enhance NAT \cite{li2019hint}.
(8) \textbf{Flowseq-large} which integrates generative flow into NAT \cite{ma2019flowseq}.
\\ \noindent
\textbf{Model Configurations} \, We implement our approach on NAT-Base.
We closely follow the settings of \cite{gu2017non, lee2018deterministic}.
For all experiments, we use the \textit{base} configurations ($d_{model}$=512, $d_{hidden}$=2048, $n_{layer}$=6, $n_{head}$=8). 
In the main experiments, we set the number of iterations during both training ($K_{train}$) and decoding ($K_{dec}$) to 5, $\alpha$ to 0.1 and $\beta$ to 0.5. 
\\ \noindent
\textbf{Training} \,
We train the AT teacher with the original corpus, construct the distillation corpus with the sequence-level knowledge distillation \cite{kim2016sequence}, and train NAT models with the distillation corpus.
We choose Adam \cite{kingma2014adam} as the optimizer.
Each mini-batch contains 64k tokens. 
During pre-training, we stop training when the number of training steps exceeds 400k, 180k for WMT14 En$\leftrightarrow$De, WMT16 En$\leftrightarrow$Ro, respectively. 
The learning rate warms up to 5e-4 in the first 10K steps and then decays under the inverse square-root schedule. 
During fine-tuning, we fix the learning rate to 1e-5 and stop training when the number of training steps exceeds 1k.
We run all experiments on 4 Nvidia RTX 2080Ti GPUs and choose the best checkpoint by BLEU points.
\\ \noindent
\textbf{Inference} \, 
During inference, we employ length parallel decoding (LPD) which generates several candidates in parallel and re-ranks with the AT teacher \cite{gu2017non}. 
During post-process, we remove tokens generated repeatedly.
\section{Results and Analysis}
\subsection{Main Results}
Table \ref{tab:main-results} shows our main results on WMT14 En$\leftrightarrow$De and WMT16 En$\leftrightarrow$Ro datasets. 
Coverage-NAT achieves strong BLEU improvements over NAT-Base by +2.75, +2.34, +1.27 and +1.33 BLEU on WMT14 En$\rightarrow$De, WMT14 De$\rightarrow$En, WMT16 En$\rightarrow$Ro and WMT16 Ro$\rightarrow$En, respectively. 
Moreover, Coverage-NAT surpasses most NAT baselines on BLEU scores with a competitive speedup over Transformer (10.04$\times$).
It is worth mentioning that Coverage-NAT achieves higher improvements on WMT14 dataset than WMT16 dataset.
The reason may be that it is harder for NAT to align to long source sentences which are more frequent in WMT14 dataset, and Coverage-NAT exactly models the coverage information and hence leads to higher improvements on WMT14 dataset.
With the LPD of 9 candidates, Coverage-NAT achieves +5.10, +5.43, +2.91 and +1.56 BLEU of increases than NAT-FT (rescoring 10) on WMT14 En$\rightarrow$De, WMT14 De$\rightarrow$En, WMT16 En$\rightarrow$Ro and WMT16 Ro$\rightarrow$En respectively, which is near the performance of Transformer but 5.22$\times$ faster than Transformer on WMT16 En$\leftrightarrow$Ro tasks.
\subsection{Ablation Study}
\begin{table}[t!]
    \caption{Ablation study on WMT16 En$\rightarrow$Ro development set. We show the effects of different iterations during training ($K_{train}$) and different SCA weights ($\beta$).}
    \centering
    \resizebox{\columnwidth}{!}{
        \begin{tabular}{l|c|c|c|c}
            \toprule
            \multicolumn{1}{l|}{\multirow{2}[2]{*}{Model}} & \multirow{2}[2]{*}{$K_{train}$} & \multirow{2}[2]{*}{$\beta$} & \multirow{2}[2]{*}{BLEU (\%)} & \multicolumn{1}{c}{Latency} \\
            & & & & \multicolumn{1}{c}{(ms)} \\
            \midrule
            NAT-Base & /     & /     & 28.96 & 20.51 \\
            \midrule
            \multirow{6}[1]{*}{Coverage-NAT} & 1     & /     & 28.77 (-0.19) & 20.82 \\
            & 3     & /     & 29.20 (+0.24) & 24.84 \\
            & 5     & /     & 29.33 (+0.37) & 28.34 \\
            & 10     & /     & 29.45 (+0.49) & 40.80 \\
            & 15     & /     & 29.61 (+0.65) & 48.90 \\
            & 5     & 0 & 29.59 (+0.63) & 28.34 \\
            & 5     & 1 & 30.03 (+1.07) & 28.34 \\
            & 5     & 0.5 & 30.16 (+1.20) & 28.34 \\
            \bottomrule
    \end{tabular}}%
    \label{tab:ablation-study}%
\end{table}%
As Table \ref{tab:ablation-study} shows, we conduct the ablation study on WMT16 En$\rightarrow$Ro development set to show the effectiveness of each module in Coverage-NAT.
First, we evaluate the effects of different iterations during training ($K_{train}$).
In ablation study, $K_{train}$ and $K_{dec}$ are the same.
With the increase of $K_{train}$, the BLEU firstly drops slightly then increases continuously.
It demonstrates that the TCIR needs enough iterations to learn effective source exploitation with the help of token-level coverage information.
Because of the trade-off between performance and latency, we set $K_{train}$ to 5 in our main experiments for an equilibrium although $K_{train}=15$ brings more gains.
Second, we evaluate the effects of different SCA weights.
When we set $\beta$ to 0, 1 and 0.5, the BLEU increases by +0.63, +1.07 and +1.20, respectively.
It proves that a moderate $\beta$ is important for the effects of SCA.
\subsection{Effects of Iterations During Decoding}
\begin{figure}[t!]
    \centering
    \includegraphics[width=0.9\linewidth]{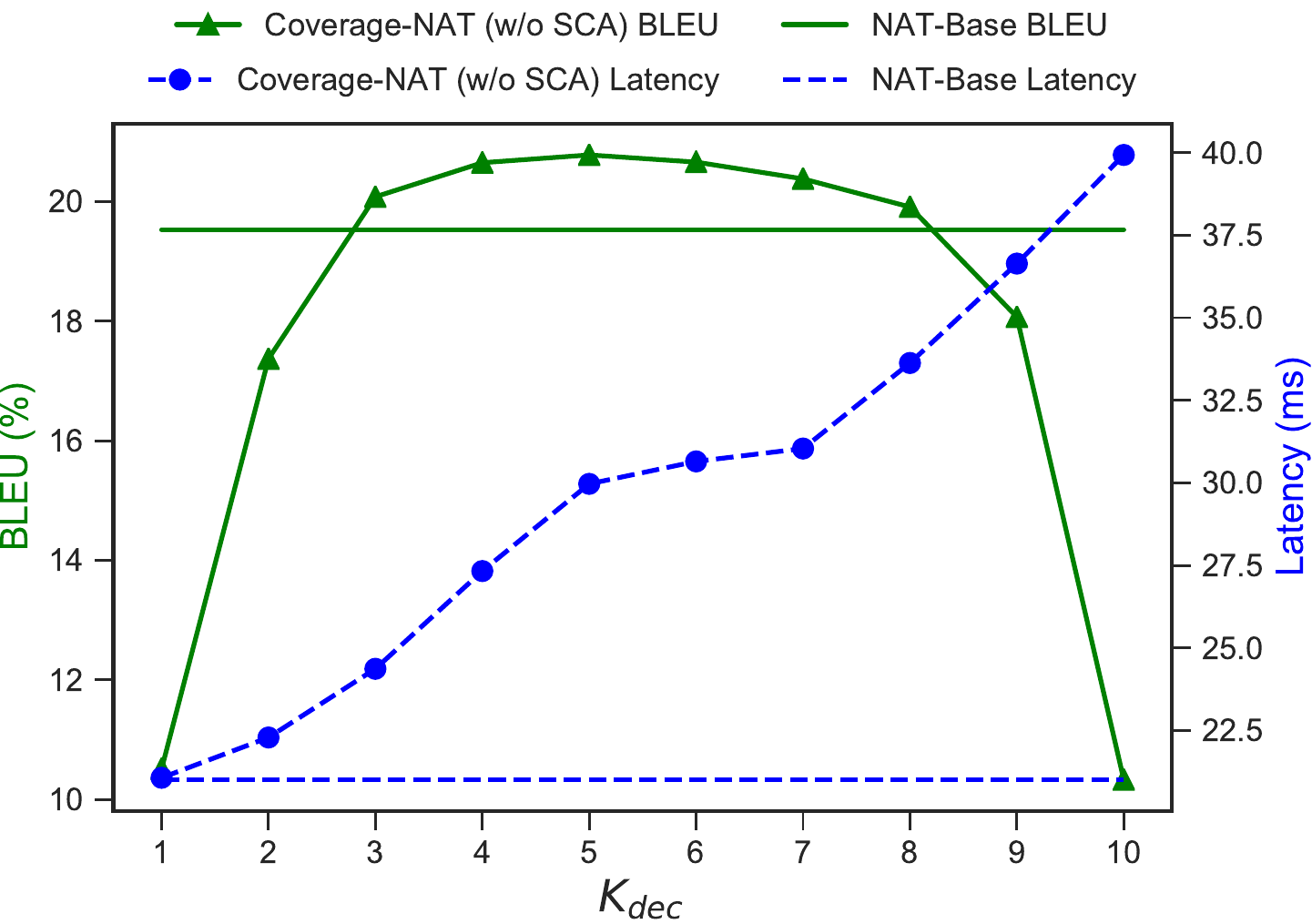}
    \caption{The BLEU and the latency with different iterations during decoding ($K_{dec}$) on WMT14 En$\rightarrow$De development set. The results are from Coverage-NAT (\textit{w/o} SCA, $K_{train}=5$). It shows that our model reaches the maximum BLEU when $K_{dec}=K_{train}$, and achieves almost the maximum improvement when $K_{dec}=4$ which indicates our model can work well with less iterations and lower latency.}
    \label{fig:deciter}
\end{figure}
\begin{table}[t]
    \caption{The ratio of repeated tokens (\%) for translations on the WMT14 En$\rightarrow$De development set. It proves that \textit{over-translation} errors grow as the sentence becomes longer and the SCA achieves higher improvements than the TCIR.}
    \centering
    \resizebox{\columnwidth}{!}{
        \begin{tabular}{lccc}
            \toprule
            Model & Short & Long & All \\
            \midrule
            NAT-Base & 5.47 & 12.05 & 10.23 \\
            Coverage-NAT & 3.31 (-2.16) & 7.52 (-4.53) & 6.36 (-3.87) \\
            ~~~--~SCA & 4.82 (-0.65) & 10.21 (-1.84) & 8.72 (-1.51) \\
            \bottomrule
    \end{tabular}}
    \label{tab:repeated-token}
\end{table}
As Figure \ref{fig:deciter} shows, we use Coverage-NAT (\textit{w/o} SCA, $K_{train}=5$) to decode the WMT14 En$\rightarrow$De development set with different iterations during decoding ($K_{dec}$). 
As $K_{dec}$ grows, the BLEU increases and gets the maximum value near $K_{train}$ while the latency increases continuously. 
The reason that the performance doesn't increase continuously may be that our model are trained to provide the best coverage at $K_{train}$ and this consistency will be broken if $K_{dec}$ stays away from $K_{train}$.
Our model surpasses NAT-Base when $K_{dec}=3$ and achieves almost the maximum BLEU when $K_{dec}=4$, which indicates our model can work well with less iterations and lower latency than $K_{train}$ during decoding.
\subsection{Analysis of Repeated Tokens}
\label{sec:repeated-tokens}
We study the ratio of repeated tokens during post-process to see how our methods improve the translation quality. We split the WMT14 En$\rightarrow$De development set equally into the short part and the long part according to the sentence length. 
Then, we translate them using NAT models and score the ratio of repeated tokens during post-process.
As shown in Table \ref{tab:repeated-token}, NAT-base suffers from \textit{over-translation} errors especially on the long sentences, which is in line with \cite{shao2019minimizing}.
As the sentence length grows, it becomes harder for NAT to align to source, which leads to more translation errors and eventually damages the translation quality.
However, the ratio of repeated tokens reduces obviously in Coverage-NAT by 2.16\% and 4.53\% on short and long sentences, which proves that our model achieves higher improvements as the sentence length becomes longer.
Moreover, we remove the SCA from Coverage-NAT and compare it with NAT-Base. It leads to 0.65\% and 1.84\% decreases of repeated tokens on short and long sentences, respectively.
The results prove that both token-level and sentence-level coverage can reduce \textit{over-translation} errors and sentence-level coverage contributes more than token-level coverage.

\begin{figure}[t!]
    \centering
    \includegraphics[width=0.9\linewidth]{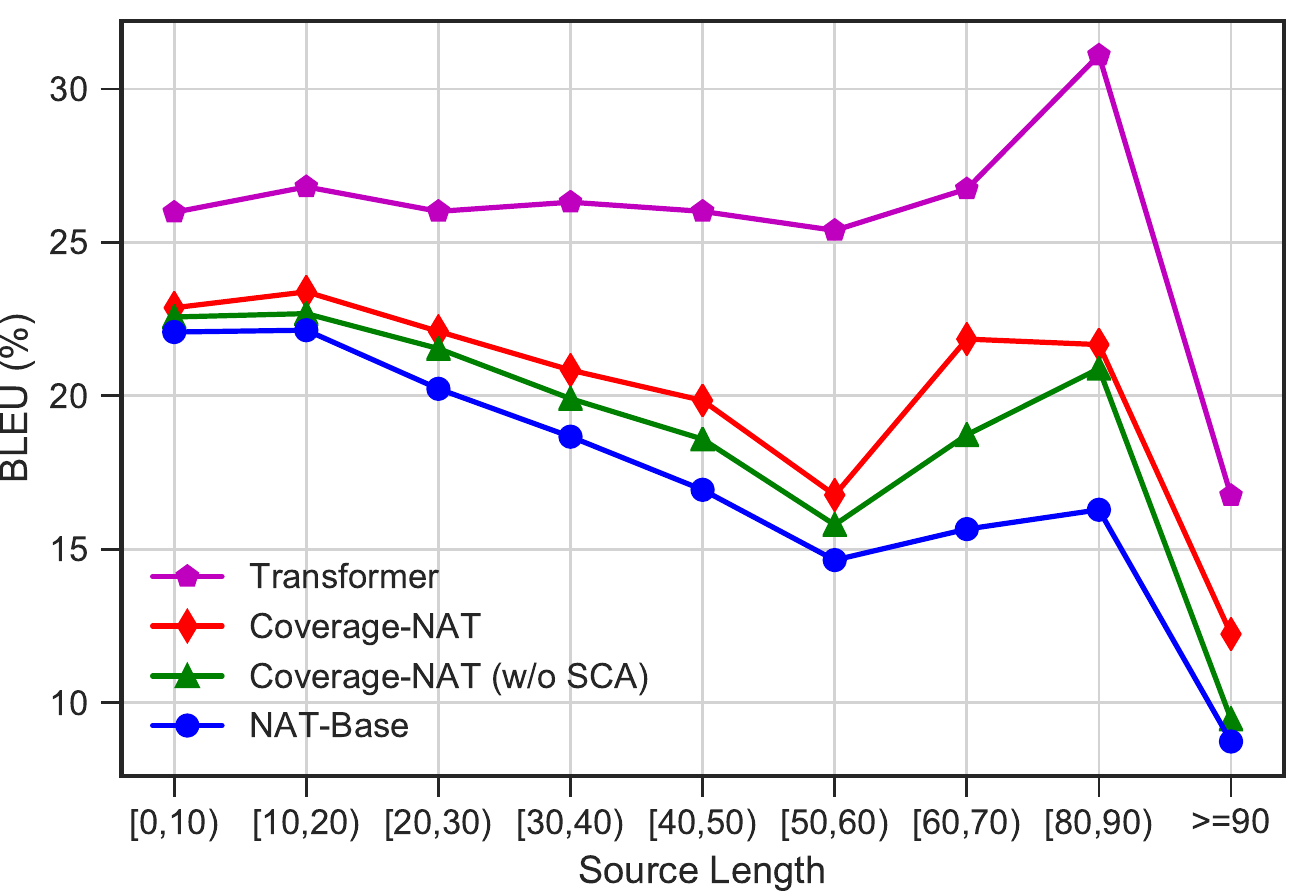}
    \caption{The BLEU with different source lengths on the WMT14 En$\rightarrow$De development set. Compared to NAT-Base, Coverage-NAT is closer to Transformer with a slighter decline in ``[10,50)" and a greater increase in ``[50,80)".}
    \label{fig:bleu-with-srclen}
\end{figure}
\begin{table}[t]
    \caption{Subjective evaluation of \textit{under-translation} errors. The higher the metric, the less the errors.}
    \centering
    \resizebox{0.95\columnwidth}{!}{
        \begin{tabular}{lc}
            \toprule
            Model & Under-Translation Metric \\ 
            \midrule
            NAT-Base & 58.9\% \\ 
            Coverage-NAT (\textit{w/o} SCA) & 67.1\% \\ 
            Coverage-NAT & 70.6\% \\ 
            \bottomrule 
    \end{tabular}}
    \label{tab:subj-eval}
\end{table}
\begin{table*}[t!]
    \caption{Translation examples from the WMT14 De$\rightarrow$En test set. We use underlines to represent \uline{\textit{over-translation}} errors and wavy lines to represent \uwave{\textit{under-translation}} errors. Note that the text in brackets is added to represent the missing parts other than the exact translation. Our model can alleviate those errors and generate better translations.}
    \centering
    \resizebox{\textwidth}{!}{
        \begin{tabular}{l|l}
            \hline 
            \hline
            Source & Edis erklärte , Coulsons Aktivitäten bei dieser Story folgten demselben Muster wie bei anderen \bigstrut[t]\\
            & wichtigen Persönlichkeiten , etwa dem früheren Innenminister David Blunkett . \bigstrut[b]\\
            \hline
            Reference & Mr Edis said Coulson \&apos;s involvement in the story followed the same pattern as with other \bigstrut[t]\\
            & important men , such as former home secretary David Blunkett . \bigstrut[b]\\
            \hline
            Transformer    & Edis explained that Coulson ’ s activities in this story followed the same pattern as those of other \bigstrut[t]\\
            & important personalities , such as former Interior Minister David Blunkett . \bigstrut[b]\\
            \hline
            NAT-Base & Edis explained that Coulson ’ s activities in this story followed the same pattern as \uline{other other} \bigstrut[t]\\
            & \uwave{(important)} \uline{person personalities} such as \uline{former former} \uwave{(Inter)ior} Minister David Blunkett . \bigstrut[b]\\
            \hline
            Coverage-NAT (\textit{w/o} SCA) & Edis explained that Coulson ’ s activities in this story followed the same pattern as \uline{other other} \bigstrut[t]\\
            & important personalities , such as former Interior Minister David Blunkett . \bigstrut[b]\\
            \hline
            Coverage-NAT & Edis explained that Coulson ’ s activities in this story followed the same pattern as with other \bigstrut[t]\\
            & important personalities , such as former Interior Minister David Blunkett . \bigstrut[b]\\
            \hline 
            \hline
    \end{tabular}}%
    \vspace{-10pt}
    \label{tab:case}%
\end{table*}%
\subsection{Subjective Evaluation of Under-Translation}
We conduct a subjective evaluation to validate our improvements to \textit{under-translation} errors.
For each sentence, two human evaluators are asked to select the one with the least \textit{under-translation} errors from translations of three anonymous systems.
We evaluate a total of 300 sentences sampled randomly from the WMT14 De$\rightarrow$En test set.
Then, we calculate the \textit{under-translation} metric which is the frequency that each system is selected.
The higher the \textit{under-translation} metric, the less the \textit{under-translation} errors.
Table \ref{tab:subj-eval} shows the results.
Note that there are some tied selections so that the sum of three scores does not equal to 1.
It proves that Coverage-NAT produces less \textit{under-translation} errors than NAT-Base, and both word-level and sentence-level coverage modeling are effective in reducing them.
\subsection{Effects of Source Length}
To evaluate the effects of sentence lengths, we split the WMT14 En$\rightarrow$De development set into different buckets according to the source length and compare the BLEU performance of our model with other baselines.
Figure \ref{fig:bleu-with-srclen} shows the results.
As the sentence length grows, the performance of NAT-Base firstly drops continuously, then has a slight rise and finally declines heavily on very long sentences.
However, Transformer can keep a high performance until the sentence length becomes very long, which indicates that translating very long sentences is exactly difficult.
Compared to NAT-Base, the result of Coverage-NAT is closer to Transformer, which has a more slight decline in ``[10,50)" and a greater increase in ``[50,80)".
The result of Coverage-NAT (\textit{w/o} SCA) lies between Coverage-NAT and NAT-Base.
It demonstrates that our model generates better translations as the sentence becomes longer and both token-level and sentence-level coverage modeling are effective to improve the translation quality, in line with Table \ref{tab:repeated-token}.
\subsection{Case Study}
We present several translation examples on the WMT14 De$\rightarrow$En test set in Table \ref{tab:case}, including the source, the reference and the translations given by Transformer, NAT-Base, Coverage-NAT (\textit{w/o} SCA) and Coverage-NAT.
As it shows, NAT-Base suffers severe \textit{over-translation} (e.g. ``other other") and \textit{under-translation} (e.g. ``important") errors. 
However, Our Coverage-NAT can alleviate these errors and generate higher quality translations.
The differences between Coverage-NAT (\textit{w/o} SCA) and Coverage-NAT further prove the effectiveness of sentence-level coverage.
\section{Related Work}
Gu et al. \cite{gu2017non} introduce non-autoregressive neural machine translation to accelerate inference.
Recent studies have tried to bridge the gap between non-autoregressive models and autoregressive models.
Lee et al. \cite{lee2018deterministic} refine the translation iteratively where the outputs of decoder are fed back as inputs in the next iteration. 
Ghazvininejad et al. \cite{ghazvininejad2019mask} propose a mask-predict method to generate translations in several decoding iterations. 
\cite{shu2019latent,gu2019levenshtein} are based on iterative refinements, too.
These methods refine the translation by decoding iteratively while our TCIR refines the source exploitation by iteratively modeling the coverage with an individual layer.
Besides, Libovick{\`y} and Helcl \cite{libovicky2018end} introduce a connectionist temporal classification loss to guide the NAT model.
Shao et al. \cite{shao2019retrieving,shao2019minimizing} train NAT with sequence-level objectives and bag-of-ngrams loss.
Wang et al. \cite{wang2019non} propose NAT-REG with two auxiliary regularization terms to alleviate \textit{over-translation} and \textit{under-translation} errors.
Guo et al. \cite{guo2019non} enhance the input of the decoder with target-side information.
Some works introduce latent variables to guide generation \cite{kaiser2018fast,akoury2019syntactically,shu2019latent,ma2019flowseq}.
Ran et al. \cite{ran2019guiding} and Bao et al. \cite{bao2019non} consider the reordering information in decoding.
Li et al. \cite{li2019hint} use the AT to guide the NAT training.
Wang et al. \cite{wang2018semi} and Ran et al. \cite{ran2020learning} further propose semi-autoregressive methods.

To alleviate the \textit{over-translation} and \textit{under-translation} errors, coverage modeling is commonly used in SMT \cite{koehn2003statistical} and NMT \cite{tu2016modeling, mi2016coverage,li2018simple,zheng2018modeling,zheng2019dynamic}. 
Tu et al. \cite{tu2016modeling} and Mi et al. \cite{mi2016coverage} model the coverage in a step-wise manner and employ it as additional source features, which can not be applied to NAT directly.
However, our TCIR models the coverage in an iteration-wise manner and employs it as attention biases, which does not hurt the non-autoregressive generation.
Li et al. \cite{li2018simple} introduce a coverage-based feature into NMT.
Zheng et al. \cite{zheng2018modeling,zheng2019dynamic} explicitly model the translated and untranslated  contents.

Agreement-based learning in NMT has been explored in \cite{liu2016agreement,cheng2016agree,zhang2019regularizing,al2019consistency}.
Yang et al. \cite{yang2019sentence} propose a sentence-level agreement between source and reference while our SCA considers the sentence-level agreement between source and translation.
Besides, Tu et al. \cite{tu2017neural} propose to reconstruct source from translation, which improves the translation adequacy but can not model the sentence-level coverage directly.
\section{Conclusion}
In this paper, we propose a novel Coverage-NAT which explicitly models the coverage information at token level and sentence level to alleviate the \textit{over-translation} and \textit{under-translation} errors in NAT.
Experimental results show that our model can achieve strong BLEU improvements with competitive speedup over the baseline system.
The analysis further prove that it can effectively reduce translation errors.

\normalem \bibliography{ijcnn2021}
\bibliographystyle{IEEEtran}

\end{document}